\documentclass{llncs}
\usepackage{amssymb}
\usepackage{makeidx}  
\usepackage{url}     
\usepackage{amsmath}
\usepackage{nccmath}
\usepackage{amsfonts} 
\usepackage{graphicx}
\usepackage{multirow}
\usepackage{bm}

\usepackage{array,booktabs,arydshln,xcolor}
\usepackage{cite}
\usepackage{floatrow}
\begin{document}
\frontmatter          
\pagestyle{headings}  
\mainmatter              
\title{Let's agree to disagree: learning highly debatable multirater labelling.}%
 
\author{Carole H. Sudre\inst{1}${}^{,}$\inst{2}${}^{,}$\inst{3}
\and Beatriz Gomez Anson\inst{4} 
\and Silvia Ingala\inst{5}
\and Chris D Lane\inst{2} 
\and Daniel Jimenez\inst{2} 
\and Lukas Haider\inst{6}
\and Thomas Varsavsky\inst{1}${}^{,}$\inst{3}
\and Ryutaro Tanno \inst{3}
\and Lorna Smith\inst{7}
\and S{\'e}bastien Ourselin\inst{1}
\and Rolf H J{\"a}ger\inst{8}
\and M. Jorge Cardoso\inst{1}${}^{,}$\inst{2}${}^{,}$\inst{3}}

\institute{
 School of Biomedical Engineering and Imaging Sciences, KCL, UK
\and Dementia Research Centre, UCL Institute of Neurology, UK
\and Department of Medical Physics and Biomedical Engineering, UCL, UK\\
\and  Santa Creu i Sant Pau Hospital, Universitat Autonòma Barcelona,  Spain
\and Vrije University Medical Centre Amsterdam, The Netherlands
\and  Queen Square Multiple Sclerosis Centre, UCL Institute of Neurology, London, UK
\and  Cardiometabolic Phenotyping Group, Institute of Cardiovascular Science, UCL,UK
\and Brain Repair and Rehabilitation Group, Institute of Neurology, UCL, London, UK}

 \index{Sudre, Carole H.}
\index{Gomez Anson, Beatriz}
\index{Ingala, Silvia}
\index{Lane, Chris D.}
\index{Jimenez, Daniel}
\index{Haider, Lukas}
\index{Varsavsky-Aisemberg, Thomas}
\index{Tanno, Ryutaro}
\index{Smith, Lorna}
\index{Ourselin, S{\'e}bastien}
\index{J{\"a}ger, Rolf H.}
 \index{Cardoso, Manuel Jorge}
 
\authorrunning{C.H. Sudre et al.}

 
\maketitle
\begin{abstract}
Classification and differentiation of small pathological objects may greatly vary among human raters due to differences in training, expertise and their consistency over time. In a radiological setting, objects commonly have high within-class appearance variability whilst sharing certain characteristics across different classes, making their distinction even more difficult. As an example, markers of cerebral small vessel disease, such as enlarged perivascular spaces (EPVS) and lacunes, can be very varied in their appearance while exhibiting high inter-class similarity, making this task highly challenging for human raters. In this work, we investigate joint models of individual rater behaviour and multi-rater consensus in a deep learning setting, and apply it to a brain lesion object-detection task. Results show that jointly modelling both individual and consensus estimates leads to significant improvements in performance when compared to directly predicting consensus labels, while also allowing the characterization of human-rater consistency.
\end{abstract}
\section{Introduction}
\label{sec:intro}

Detection and differentiation between types of pathological objects is a core problem of medical image analysis, generally requiring costly expert labelling. Disagreement between raters can be a result of differences in radiological training schools, rater competence, and sample appearance, among others. This problem is often exacerbated by changes in rater performance caused by retraining or observational bias.

Due to the variability in shape and intensity signatures observed across the full spectrum of lesions, even the most trained raters can present a high inter-rater variability. In such cases, finding a majority voting consensus classification is the most common strategy.  
 
When classifying objects into multiple classes, it is often more complex to separate all object types directly, than it is to first detect all pathological objects followed by their classification, as some class decision boundaries are easier than others.
This sequential detection/classification problem is, for instance, present in the context of age-related vascular changes in which macroscopic alterations can be observed on structural MR images. Among these observed changes, small elements such as enlarged perivascular spaces (EPVS) and lacunes are observed on similar image sequences \cite{Wardlaw2013SVDStandards}. EPVS, often associated with concomitant neuropathology and deleterious clinical outcome \cite{Ramirez2016}, appear as fluid filled structures with a linear shape. However, because of their limited size ($<$10mm$^3$) and highly variable appearance, EPVS instances are often confused with other concomitant lesions, such as lacunes.
Because of this intrinsic uncertainty, the labelling of these small vascular lesions can be seen as a four-class problem, with classes `EPVS', `Lacune', `Undetermined', and `Nothing'. This classification problem suffers from two concomitant issues: a) class imbalance with a 100:1 ratio between EPVS and lacunes, and b) noisy labelling as a result of rater disagreement. Consensus labels are also problematic in this setting, as rater behaviour is non-random and samples are not truly independent. 

In this work, we build on a previously described 3-dimensional multirater Regional Convolutional Neural Network (RCNN) model, used here as a lacune and EPVS object detection system. However, rather than only learning the consensus value or a single rater, we propose to jointly learn the consensus majority voting, the associated probability for each class, and each individual rater decision, so as to appropriately model highly debatable label predictions. 

\section{Related work and problem specificities}
Many of the recent publications on classification with noisy labels assume independence between samples and noise, and a constant mislabelling probability \cite{Li2017}, which does not hold in the case of difficulty induced variability and rater shift.
Strategies for classification in the presence of noise include sample re-weighting (importance reweighting) or curriculum-based sample selection\cite{Bouguelia2018,Jiang2017}. Other approaches, normally classed as label/classifier fusion, disentangle rater and label uncertainty either by iteratively favouring raters that agree with the consensus \cite{Warfield2004}, or by reducing sample correlation to construct a balanced classifier \cite{HongzhiWang2013}. Lastly, the relationships between rater behaviour can also be learned through their confusion matrices \cite{Tanno}. Notably, most of these works focus on the problem of classification, where balanced sampling strategies can be employed, something that is not possible in a joint detection/classification model. In this work, we argue that combining majority voting predictions while learning individual rater behaviour allows for a better model of rater consistency and sample uncertainty. 

\begin{figure}
    \centering
    \includegraphics[width=0.95\textwidth]{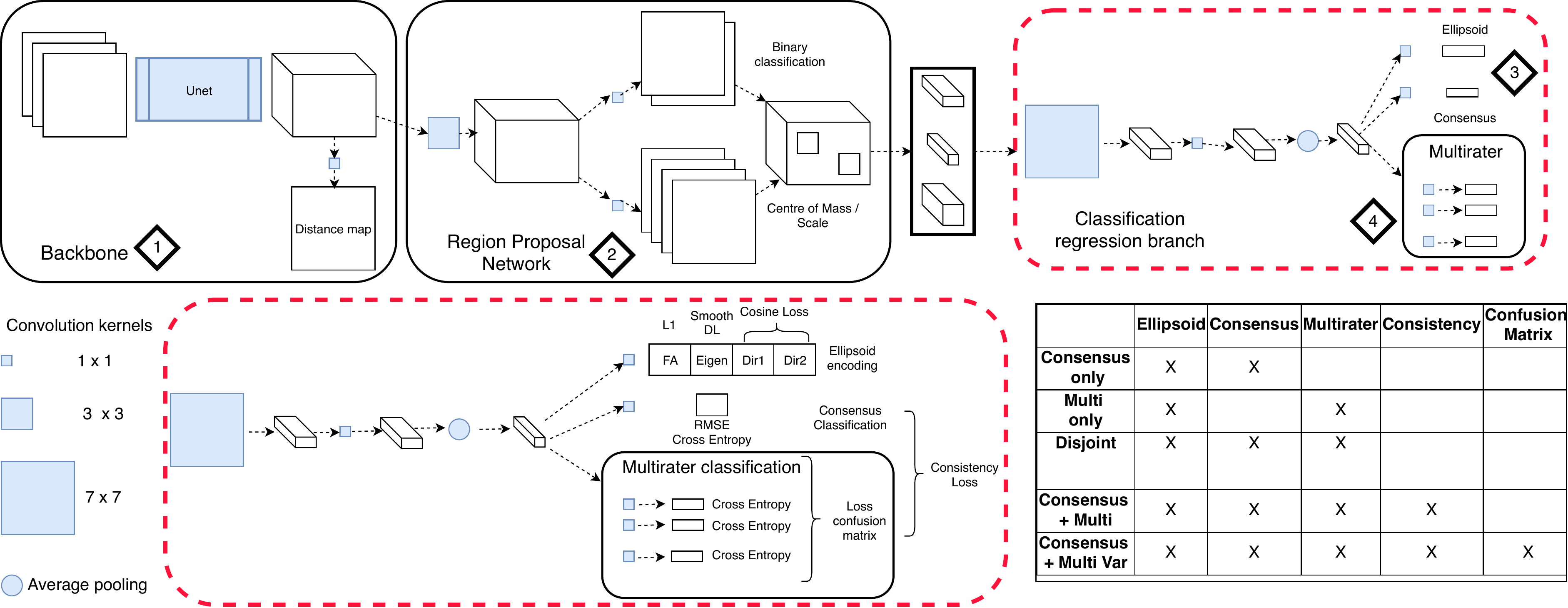}
    \caption{Detection and multirater classification architecture framework. In this work focus is put on the classification branch (red dashed) further detailed on the second row along with the description of the different training regimes.}
    \label{fig:NetworkTraining}
\end{figure}

\section{Methods}
\label{sec:methods}
\subsection{Network architecture}
The multirater 3D RCNN framework presented in \cite{Sudre2019} is composed of four stages: 1) a backbone network learning the features using as target a distance map to the objects of interest; 2) a region proposal network (RPN) regressing the location of candidate centres of mass of target objects together with their spatial scale; 3) patches from the RPN representation are fed into a two layer network in order to regress the average object classification and object shape; 4) a multibranch fully connected layer is used to model the behaviour of each rater. (see Fig. \ref{fig:NetworkTraining}.) 
\paragraph{Shape encoding}
Instead of modelling each object by its encompassing cuboid \cite{Sudre2019}, the shape of each candidate object is encoded as a four-parameter simplified encompassing ellipsoid, namely using the largest eigenvalue, the two first components of the associated eigenvector and the value of fractional anisotropy of the associated tensor. 
\subsection{Multi-rater classification}
The classification of the candidate objects is defined as a four-class problem, i.e. EPVS, Lacune, Undetermined, Nothing. 
From a human-rater point of view, the classification can be seen in two different ways: a multi-rater consensus, here modeled as the probability of a class to be chosen among the six raters computed as the average rating, and a rater-specific categorical label.
\paragraph{Consensus average classification}
When modelling the consensus/average rater, the training can be performed using either a hard or a probabilistic classification, or a combination of both. Note, here, the hard classification corresponds simply to the majority voting categorical consensus, while the continuous probability encodes the uncertainty over the final classification. As a consequence, a cross-entropy loss is used to learn the consensus, while a root mean square error loss is used over the resulting class probabilities.
\paragraph{Independent rater modelling}
In the last stage of classification, a cross-entropy loss is used to learn each rater label independently. Inter-rater behaviour can be enforced through a variability loss ($L_{var}$) penalizing the difference between the effective and predicted probabilistic confusion matrices. Noting $C$ (resp. $\widehat{C}$) the observed (resp. predicted) confusion matrix, $L_{var}=\sum_{(i,j)} \vert C_{i,j} - \widehat{C_{i,j}}\vert$

Ideally, we would also like to have consistency between the predicted group consensus and the consensus of individual prediction. In order to achieve this, the following consistency loss $L_{cons}$ is introduced:
\[L_{cons} = \sqrt{\sum_{k=1}^{K}\widehat{p_k} - \dfrac{1}{R}\sum_{r=1}^{R}\widehat{p_{kr}}} \]
\noindent with $\widehat{p_k}$ denoting the predicted consensus probability, and $\widehat{p_{kr}}$ denoting the predicted probability given by rater $r$ for class $k$.
\paragraph{Compensating for inter-rater variability and enhancing individual rater characteristics}
The EPVS labelling problem is highly variable in terms of rater agreement; sometimes all raters agree with each other, while other times raters converge to completely different decisions. As a consequence, when predicting the group consensus, we have enforced consensus learning from samples of high agreement. To this effect, sample importances were downweighted according to their observed variability, here expressed as $var=1-\sum_{k=1}^{K}p_k^2$, where $p_k$ is the observed classification probability for class $k$. The sample is then weighted by $\exp\left(-var\right)$. Conversely, when modelling individual raters, and in order to learn rater-specific behaviours, we promote samples for which the individual rater disagrees with the consensus. This is achieved by weighting each rater-sample combination by the inverse of its contribution to the consensus ($1/p_{k^{r}}$), where $p_{k^{r}}$ is the observed probability for the sample to be classified as $k$ if rater $r$ labels it as $k$.
\section{Data and experiments}
\subsection{Data}
\label{sec:data}
16 subjects that were part of a large tri-ethnic cohort investigating the relationship between cardiovascular risk factors and brain health \cite{Tillin2012SABRECohort} were chosen due to their elevated vascular burden. 4147 EPVS and lacunes were manually segmented using jointly 1mm3 structural MR sequences (T1, T2, FLAIR) using ITKSnap \footnote{\url{http://www.itksnap.org/pmwiki/pmwiki.php?n=Main.HomePage}}. Individual segmented lesions, defined using connected components, were then classified by six trained raters using an in house dedicated viewer. Only objects bigger than 5 voxels were used in this study, resulting in a database of 2202 elements. 14 subjects were used for training and two subjects for testing. The test set contained 184 objects that were all classified at least by one rater as EPVS. Inter-rater accuracy ranged from 0.47 to 0.92 with a mean of 0.72. 

\subsection{Training modes}
In order to investigate the model's ability to handle label noise, different training regimes were adopted (see Fig. \ref{fig:NetworkTraining}): 1) Training only the shape + consensus classification (Consensus only); 2) Staged training of the shape encoding followed by the independent rater multihead (Multi Only); 3) Staged training of shape and classification, followed by training the independent rater multihead (Disjoint); 4) Staged training of `shape and consensus only', followed by `multihead only` finishing by `shape, consensus and multihead`  with consistency loss (Consensus + Multi); 5) Training as in 4, with an extra loss over the confusion matrix $L_{var}$. All models were trained for 10000 iterations with a learning rate of $10^{-4}$ and using the Adam optimiser.

\section{Experiments and Results}
\subsection{Consensus probability}
\begin{figure}[tpb]
    \centering
    \includegraphics[width=0.99\textwidth]{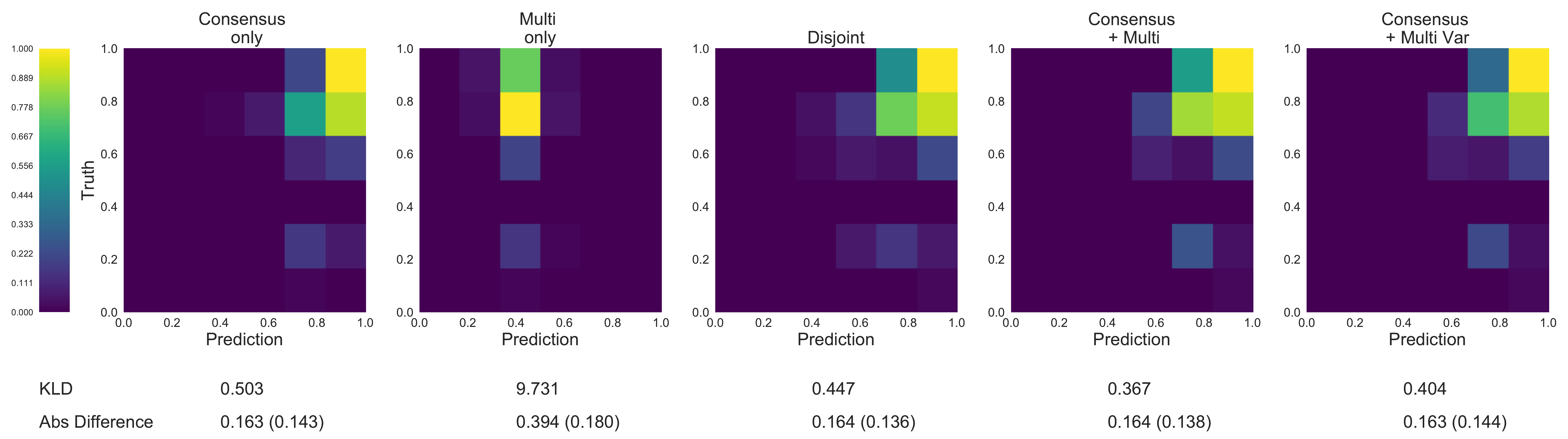}
    \caption{Comparison of EPVS probability distributions quantitatively evaluated in terms of KLD and absolute error (mean sd).}
    \label{fig:ConsensusExp}
\end{figure}

\begin{figure}[t!]
    \centering
    \includegraphics[width=0.95\textwidth,trim=0cm 6cm 0cm 0cm, clip=true]{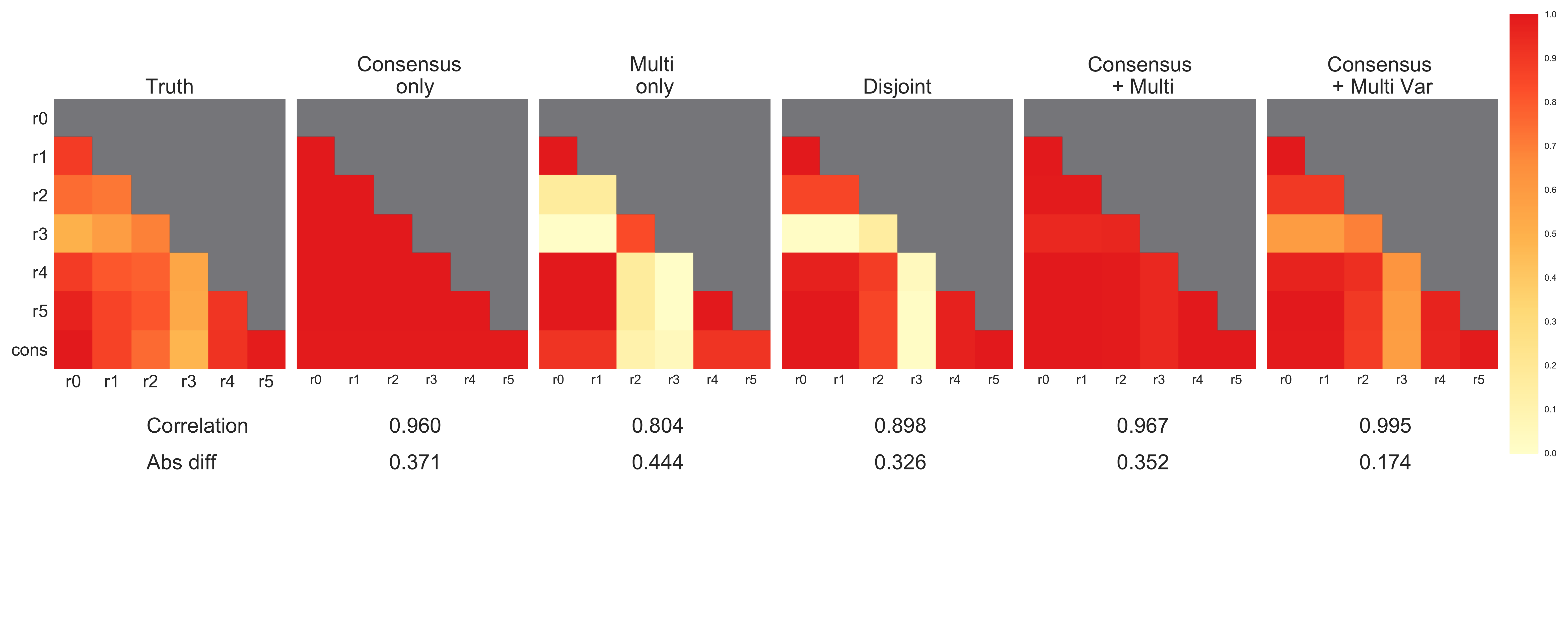}
    \caption{Pairwise agreement scores between observed rater labels compared to the agreement scores between categorized individual rater predictions for each training mode. The left most element is the target inter-rater behaviour. Pearson correlation coefficient and mean absolute difference against the target inter-rater behaviour are presented below.}
    \label{fig:MultiRaterBehaviour}
\end{figure}

As a first experiment, we investigate the ability of each training mode to appropriately predict the distribution of EPVS classification probabilities. Figure \ref{fig:ConsensusExp} presents the joint histograms of the target and predicted distributions. The resulting Kullback-Leibler Divergence (KLD) over the distributions are displayed below along with the mean absolute error in prediction. Results show that all methods explicitly learning the average consensus are able to reproduce it well.  
The ability of the different models to reproduce individual rater behaviour was evaluated by comparing predicted inter-rater agreement with observed inter-rater agreement. Figure \ref{fig:MultiRaterBehaviour} presents the pairwise agreement results between observations and between predictions, and measures of correlation and absolute difference between agreement matrices. One can note that, as expected, no inter-rater behaviour is  learnt when adopting the `Consensus only' framework. Furthermore, we observe that the inter-rater agreement learnt with both the `Multi only' and the `Disjoint' model is exacerbated compared to the truth. This rater behaviour exacerbation fades away when enforcing consistency between multi-rater consensus predictions and the consensus of individual rater predictions (i.e. `Consensus+Multi' model).

\subsection{Consistency between consensus and multirater average}
This experiment aims to test the efficacy of the loss function introduced in Section 3.2 with the aim of promoting the agreement between the multi-rater consensus labelling and the consensus of individual predictions.
\begin{figure}[b!]
    \centering
    \begin{tabular}{cc}
     \includegraphics[width=0.42\textwidth, clip=true]{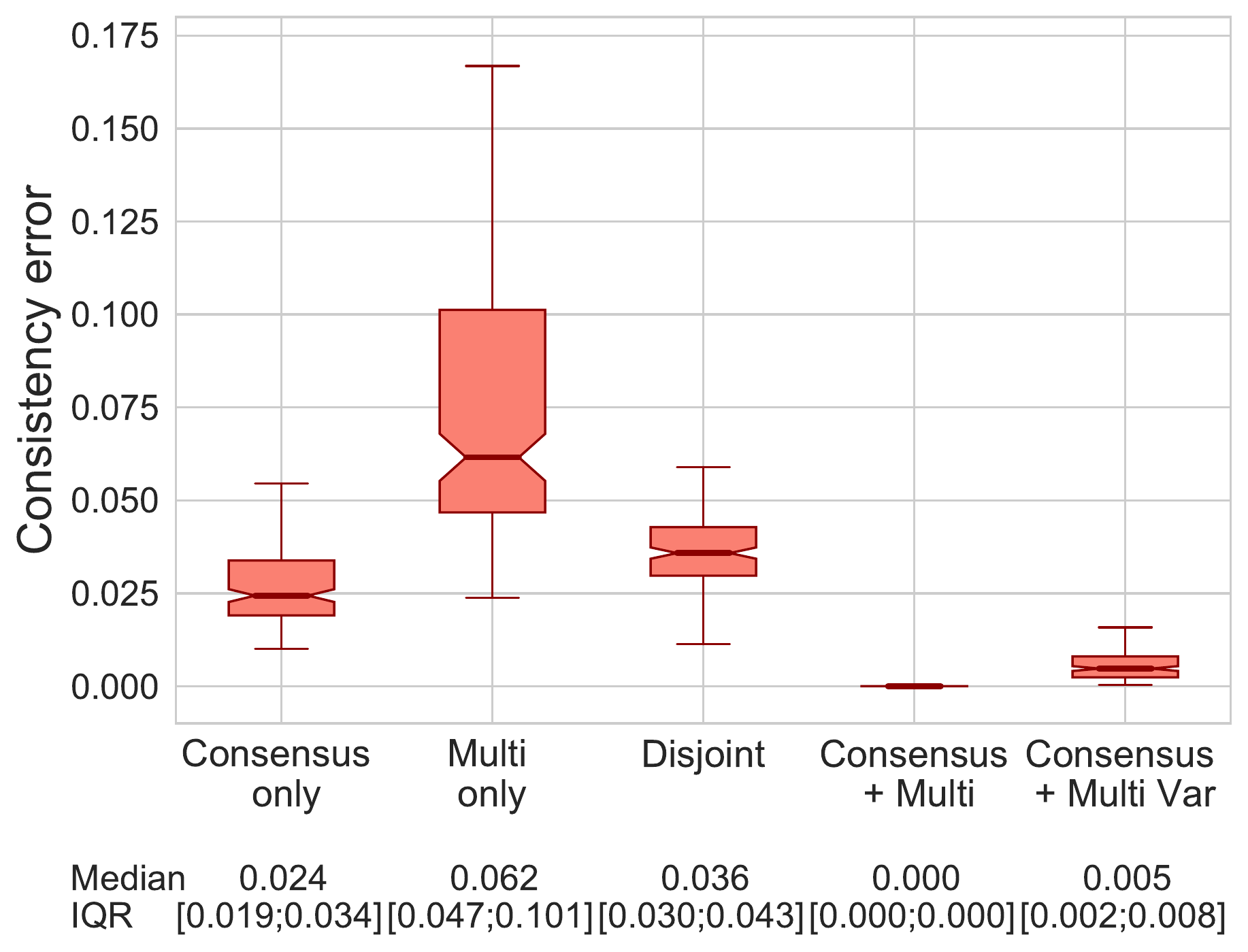}   & \includegraphics[width=0.48\textwidth]{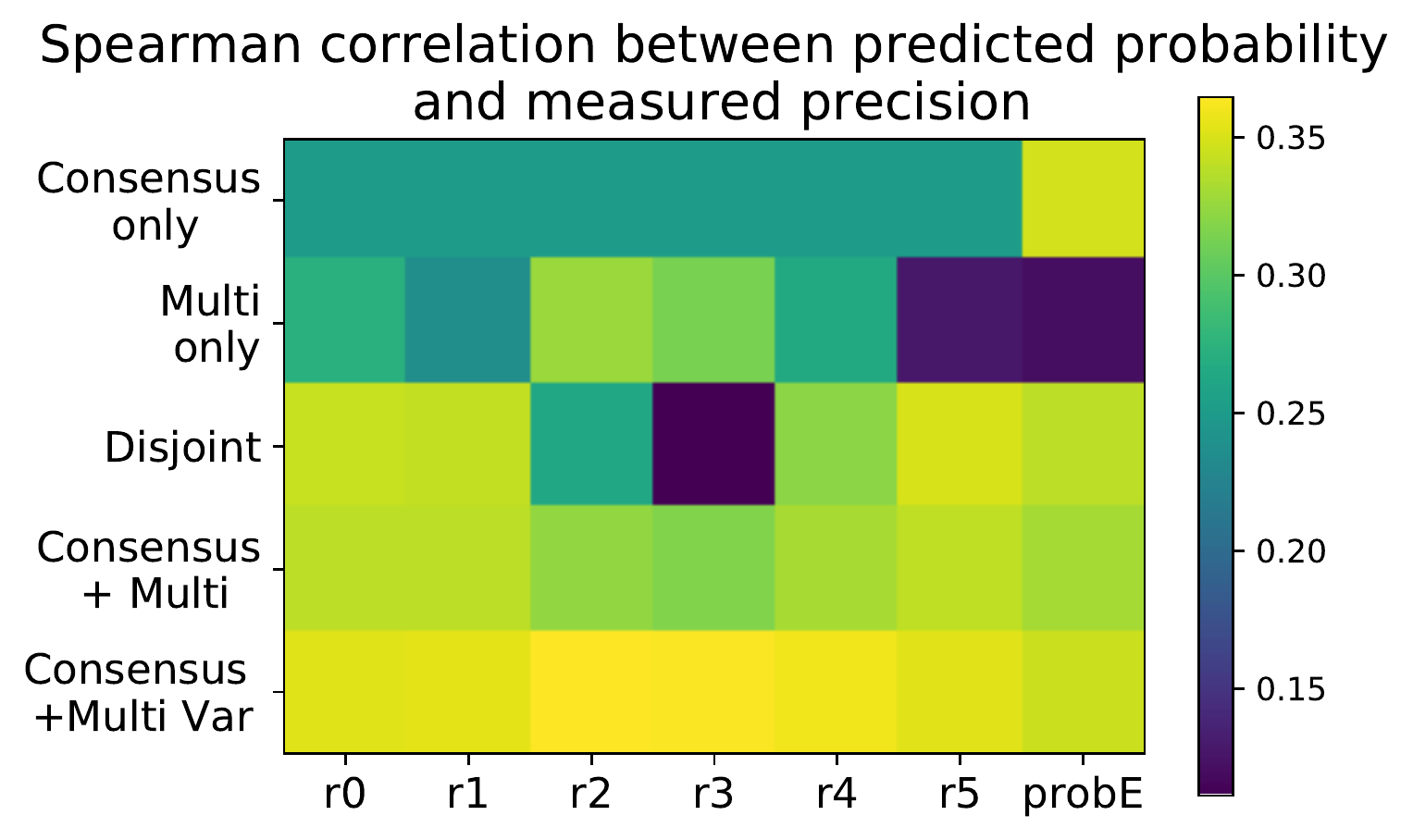}

     \end{tabular}
    \caption{Left: Boxplot of the consistency error between predicted consensus probability and average of individual raters predictions. Right: Spearman coefficient between predicted probability of classifying the element as an EPVS and measured precision ($1/var$) over multiple models. ProbE refers to the predicted probability of an EPVS for the consensus of raters. }
    \label{fig:ConsistencyVar}
\end{figure}
Figure \ref{fig:ConsistencyVar} left presents the  boxplots of the difference between the average of the predicted individual raters and the consensus prediction. Numerical results of median and interquartile range are presented below the graph. Training regimes that promote consistency between the consensus prediction and the average of independent predictions both reach, as expected, a very high level of consistency. Conversely, simpler models only optimising for independent rater predictions do not achieve a good consensus estimation. 
\subsection{Variability, disagreements and individual rater quality}
In this experiment, we would like to assess if the probabilistic predictions of each individual rater provide a good proxy for sample uncertainty (defined as the variability of individual ratings). To this end, we estimate the Spearman correlation coefficient between each individual prediction and the measured precision defined as $1/var$, displayed on Figure \ref{fig:ConsistencyVar} (right). As already noted from Figure \ref{fig:MultiRaterBehaviour}, no rater-specific information can be modeled using only the consensus. Individualized rater predictions were found to be strongly associated with overall variability, primarily when consistency losses were applied.
\begin{figure}[ptb]
    \centering
    \includegraphics[width=0.8\textwidth]{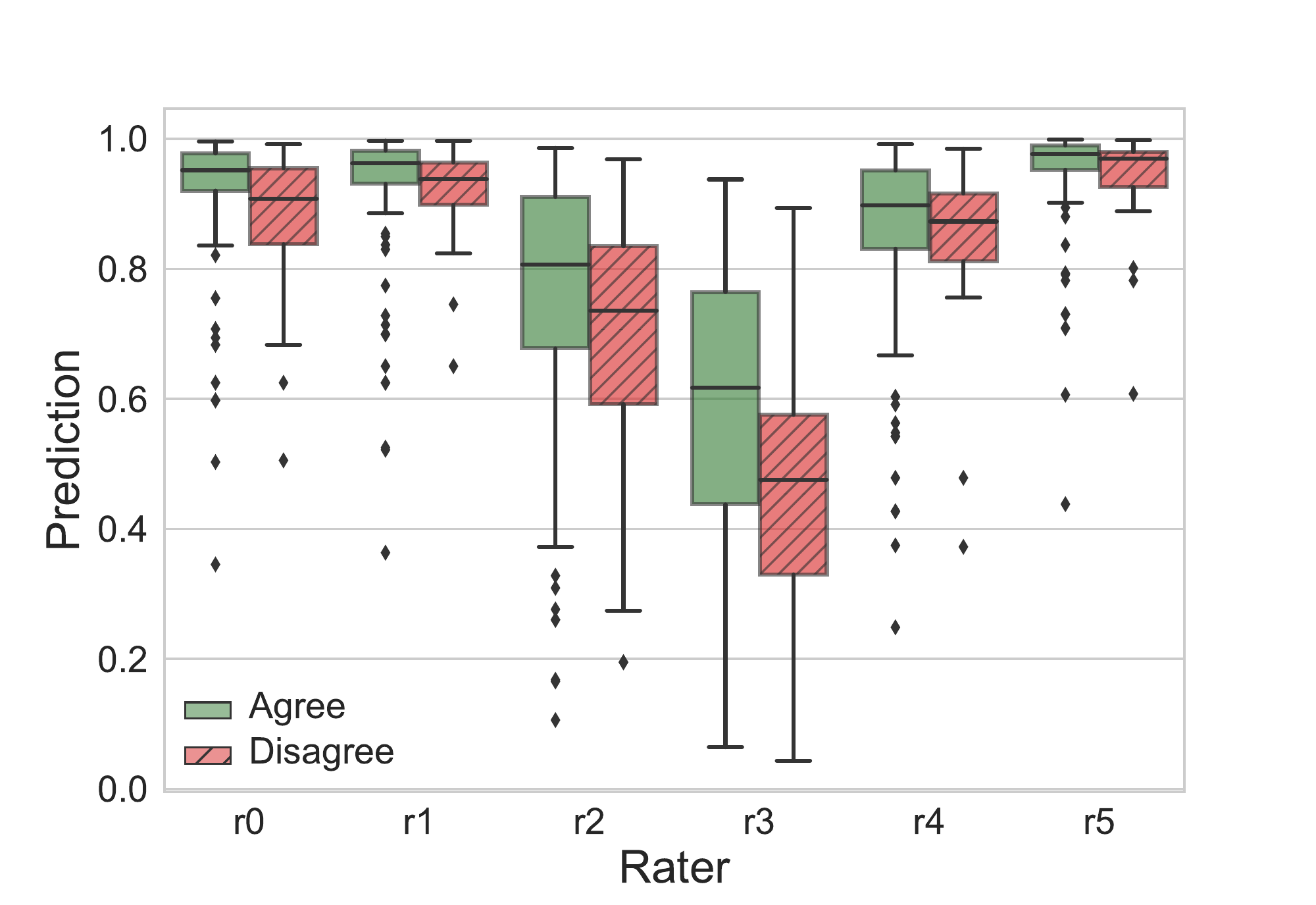}
    \caption{Boxplot of the predicted probabilities for the agreeing and disagreeing cases for each rater with the best training regime (Consensus + Multi Var).}
    \label{fig:AgreevsDisagree}
\end{figure}

\subsection{Labelling introspection}
We used the best overall model (`Consensus+Multi Var') to study the distribution of probabilistic predictions for objects whose rater classification was in agreement with the consensus versus objects where there was rater disagreement, plotted in Figure \ref{fig:AgreevsDisagree}. High prediction probabilities for individual raters were found to be a good surrogate marker rater agreement and rater consistency over training samples. Results suggest that rater 3, and to a lesser degree rater 2, displayed inconsistent labelling behaviour. Interestingly, when asked about their rating practice, both raters 2 and 3 indicated having undergone clinical retraining, possibly explaining the observed shift in their labelling. Retraining the model without these two raters resulted in an improvement in the consensus prediction, with a KLD reduced to 0.202 and a mean absolute error over the predicted probability of 0.15. This experiment suggests that one can use the proposed framework to identify not only inter-rater disagreement but also intra-rater inconsistency, and potentially correct for it.
\section{Discussion and conclusion}
\label{sec:discussion}
In this work, we investigated different training regimes in presence of noisy labelling with the aim of predicting both rater consensus and individualized predictions. We found that promoting agreement between predicted multi-rater consensus and the consensus of individualized predictions can provide good model accuracy together with the ability to introspect rater behaviour, thus not only allowing the identification of noisy labels/subjects but also assessing rater skills so as to prevent bias in large scale studies and enforce appropriate radiological training. Future work will explore the use of this information in an active learning setting and develop the accuracy of the multi-rater model estimates.

 \paragraph{Acknowledgments}We are extremely grateful to all the participants of the SABRE study, and past and present members of the SABRE team. This work was supported by an Alzheimer's Society Junior Fellowship (AS-JF-17-011), the Wellcome/EPSRC Centre for Medical Engineering [WT 203148/Z/16/Z], IMI2 grant AMYPAD [115952], the MSCA-ITN-Demo [721820], and the Wellcome Flagship Programme in High-Dimensional Neurology. The SABRE study was funded at baseline by the Medical Research Council, Diabetes UK, and the British Heart Foundation. At follow-up, the study was funded by the Wellcome Trust (067100, 37055891 and 086676/7/08/Z), the British Heart Foundation (PG/06/145, PG/08/103/ 26133, PG/12/29/29497 and CS/13/1/30327) and Diabetes UK (13/0004774). We gratefully acknowledge NVIDIA corporation for the donation of a GPU Tesla K40 that was used in the preparation of this work.
\bibliographystyle{splncs04}
\bibliography{bib2}

\end{document}